\documentclass[twocolumn]{IEEEtran}

\usepackage{amsmath,amsfonts}
\usepackage{algorithmic}
\usepackage{algorithm}
\usepackage{array}

\usepackage[space]{cite}
\usepackage{amssymb}
\usepackage{graphicx}
\usepackage{textcomp}
\usepackage{xcolor}
\usepackage{caption}
\usepackage{amsthm,bm} %
\usepackage{multirow}
\usepackage{array}
\usepackage{graphicx}
\usepackage{array,multirow}
\usepackage{booktabs}
\usepackage{caption}
\usepackage{subcaption}

\usepackage{stfloats}
\usepackage{url}
\usepackage{verbatim}

\begin{document}

\title{Significance of Skeleton-based Features in Virtual Try-On\\
}

\author{Debapriya~Roy,
        Sanchayan~Santra,
        Diganta~Mukherjee,
        and~Bhabatosh~Chanda
        
\IEEEcompsocitemizethanks{
\IEEEcompsocthanksitem D. Roy is with  the dept. of Institute of Advancing Intelligence at the TCG Centres for Research and Education in Science and Technology, Kolkata, India.\protect\\
E-mail: debapriyakundu1@gmail.com
\IEEEcompsocthanksitem Dr. S. Santra is with the Institute for Datability Science, Osaka University, Osaka, Japan.\protect\\
E-mail: sanchayan.santra@gmail.com
\IEEEcompsocthanksitem Prof. Mukherjee is with the Indian Statistical Institute, kolkata, India.\protect\\
E-mail: diganta@isical.ac.in
\IEEEcompsocthanksitem Prof. Chanda was formerly with the Indian Statistical Institute, kolkata, India and currently he is with Indian Institute of Information Technology Kalyani, India     .\protect\\
E-mail: bchanda57@gmail.com
}%
}
\markboth{IEEE TRANSACTIONS ON EMERGING TOPICS IN COMPUTATIONAL INTELLIGENCE}%
{Roy \MakeLowercase{\textit{et al.}}: Significance of Skeleton-based Features in Virtual Try-On}
\IEEEoverridecommandlockouts
\IEEEpubid{\begin{minipage}[t]{\textwidth}\ \\[10pt]
        \centering\normalsize{0000--0000/00\$00.00~\copyright~2023 IEEE}
\end{minipage}} 
\maketitle

\begin{abstract}
The idea of \textit{Virtual Try-ON} (VTON) benefits e-retailing by giving an user the convenience of trying a clothing at the comfort of their home. In general, most of the existing VTON methods produce inconsistent results when a person posing with his arms folded i.e., bent or crossed, wants to try an outfit. The problem becomes severe in the case of long-sleeved outfits. As then, for crossed arm postures, overlap among different clothing parts might happen. The existing approaches, especially the warping-based methods employing \textit{Thin Plate Spline (TPS)} transform can not tackle such cases. To this end, we attempt a solution approach where the clothing from the source person is segmented into semantically meaningful parts and each part is warped independently to the shape of the person. To address the bending issue, we employ hand-crafted geometric features consistent with human body geometry for warping the source outfit. In addition, we propose two learning-based modules: a synthesizer network and a mask prediction network. All these together attempt to produce a photo-realistic, pose-robust VTON solution without requiring any paired training data. Comparison with some of the benchmark methods clearly establishes the effectiveness of the approach. 
\end{abstract}


\section{Introduction}
\begin{figure}[!t]
	\centering
	\includegraphics[width=0.49\textwidth]{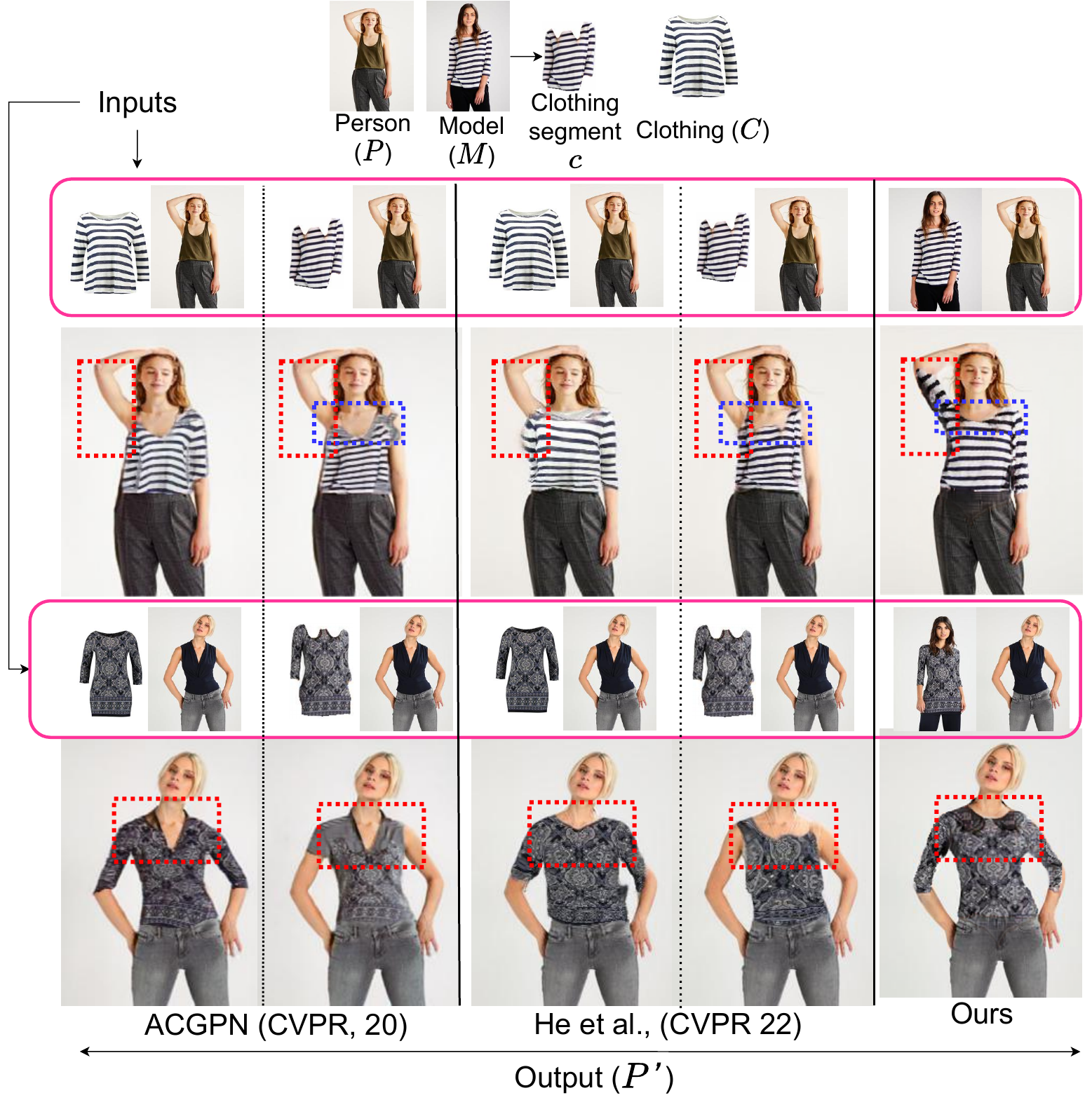}
	\captionof{figure}{Results of different methods for different input clothing, model, and person combinations. A visual comparison with some benchmark methods illustrates the efficacy of the proposed method. See }
	\label{fig: tps_problem_demo1}
\end{figure}
The online share of total global retail sales has shown a steady increase over the last few years. The fashion industry with a value of around 3 trillion US dollars is one of the major contributors in online retail sales~\footnote{https://fashinnovation.nyc/fashion-industry-statistics}. E-retailers are constantly focusing to improve consumer's shopping experience. However, buying clothing items online always includes the uncertainty of fitting and appearance. Virtual try-on (VTON) solves this problem to some extent by giving a consumer the provision to try a clothing item at the comfort of their home. Besides, for the consumers, reluctant to try a product in-store, VTON may be a great time efficient alternative solution to them.

\begin{figure}[!t]
    \centering
    \begin{minipage}{.5\textwidth}
        \centering
        \includegraphics[width=0.99\textwidth]{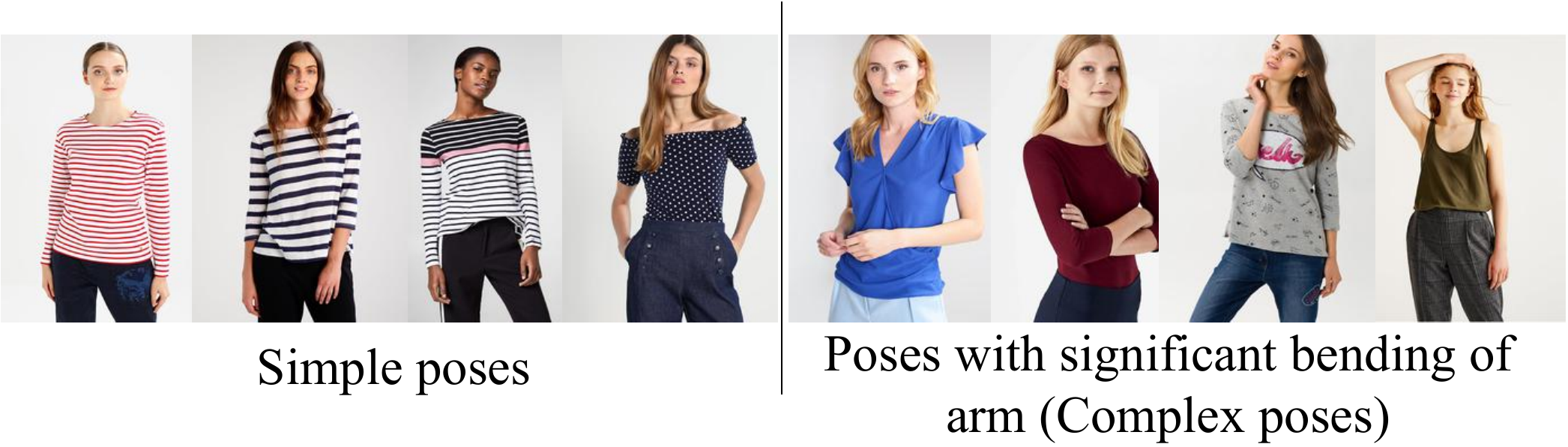}
	\captionof{figure}{Demonstration of simple and complex human poses.}
	\label{fig: demo}
    \end{minipage}%
\end{figure}

The 3D VTON methods~\cite{pons2017clothcap, sekine2014virtual, mir2020learning, bhatnagar2019multi} provides visually pleasing VTON experiences but requires 3D human body parametric models~\cite{loper2015smpl} or 3D scanned datasets. Compared to this image-based VTON approaches~\cite{cpvton, viton, lgvton, ivcnz_roy, vtnfp, multiposevton, wang2018non, fitme, fashionon, cvpr_2020_2, garmentgan, sievenet} require only two images, the garment image, and the reference person image. Most of the existing VTON methods adopt a geometric deformation-based approach. They estimates a non-linear transformation, usually the thin plate spline transformation (TPS), to deform the source clothing to fit the person. This is then combined with the person image to generate the final try-on image. Some VTON method~\cite{cpvton, multiposevton, vtnfp, cvpr_2020_2} employed a Geometric Matching Network (GMN) which is a Deep Neural Network (DNN) that predicts the TPS transform directly from the input images by leveraging the correlation between the features extracted from these images. A landmark-based image registration~\cite{image_registration} approach to compute the TPS transformation was proposed by LGVTON~\cite{lgvton}. It computes the transform using some explicitly specified source to target control point correspondences. However, both types of methods often produce inconsistent results when the person is posing with significant arm bending (see Fig.~\ref{fig: demo}). This mainly occurs due to the smoothness constraint of TPS which restricts bending in the predicted warp. In addition, the formulation of TPS models some rearrangements of control points in the image plane; therefore, not intended for applications requiring modeling of overlap or fold~\cite{bookstein1989principal} in different parts of the target warp.

Most of the approaches do not consider occlusion and pose variability in the source clothing, as they consider a separate clean garment image (see the 1$^{st}$, 3$^{rd}$ column of Fig.~\ref{fig: tps_problem_demo1}) as the source clothing. We call such methods Cloth-To-Person (C2P) methods~\cite{cpvton, viton, lgvton, ivcnz_roy, vtnfp, multiposevton, wang2018non, fitme, fashionon, cvpr_2020_2, garmentgan, sievenet, he2022style}. These methods perform poorly when the source clothing is in the form of a model wearing it. For instance, see the difference in the performance of ACGPN and He et al.~\cite{he2022style} with different clothing sources in Fig.~\ref{fig: tps_problem_demo1}. Note that the methods taking the model image as the clothing source are named as Model-To-Person (M2P) methods. M2E-TON~\cite{m2e} and Liu et al.~\cite{liu2021spatial} used the idea of body shape correspondences in M2P VTON. Compared to M2E-TON, Liu et al.'s solution can handle cases of large pose variability between model and the person and also texture-variability of clothes. However, in case of complex textures on the clothing e.g., stripes, floral, etc., it falls short in preserving details. 

In addition to the above-mentioned approaches, the problem of VTON is also explored using GAN-based~\cite{gan} attribute manipulation approaches~\cite{cagan, yildirim2019generating, men2020controllable, lewis2021tryongan}. While these methods often meet the standards of photo-realism but still fail to retain the exact texture patterns of the source clothing in the try-on output and are also limited to simple poses.

Other than the above-discussed approaches another kind is the appearance flow-based approach~\cite{chopra2021zflow, ge2021parser, han2019clothflow, he2022style}. Such methods find the cloth flow from the source to the target images. However, some of this methods~\cite {chopra2021zflow, ge2021parser, han2019clothflow} often fall short when there is a large misalignment between the garment and the person's body parts. An attempt to overcome this, by capturing the global context, is proposed in~\cite{he2022style}, but, in case of significant arm bending it shows poor performance (see Fig.~\ref{fig: tps_problem_demo1}).

Considering all these, we attempt to achieve a pose robust M2P VTON solution without requiring any paired training data, i.e., images of the same person with and without wearing the model's clothing. Unlike previous approaches we divide the source clothing into parts i.e., sleeves and torso, followed by warping each of these parts separately, before finally combining them. This approach attempts to solve the overlap issue. Moreover, it opens the flexibility of employing clothing part specific transformation. Considering the limitations of TPS we employ this for warping only that part of the clothing which is expected to undergo limited deformation, i.e., torso. However, the human arms can move in various ways causing different deformations in the sleeves most of which can not be modeled by TPS transform due to its bending constraint.

We leverage the idea of landmark registration in the current problem context~\cite{lgvton} due to its implicit consideration of human anatomical constraints; which controls undesirable deformation in the target warp. However, considering only landmark correspondences in warping the sleeves is difficult. So, inspired from \textit{field transform}~\cite{beier} we propose a hand-crafted feature-based warping technique that is consistent with human arm movements. Instead of landmark correspondences, here we consider the correspondences of the straight line segments between the consecutive landmarks. This imposes an additional constraint that is well justified with human anatomy (see Fig.~\ref{fig: all_1}(c)). 

M2P VTON methods face the inherent challenge of occlusion in the source clothing (see the clothing area near the neck occluded by the model's hair in 2$^{nd}$ row of Fig.~\ref{fig: demo}). We propose a \textit{mask prediction network (MPN)} to tackle this. The output of MPN aids in distinguishing the areas of the target clothing occluded in the source. The proposed \textit{image synthesizer network (ISN)} interpolates these occluded regions and also produces a seamless try-on image. MPN also adds to the faster computation of the target warp. 

To summarize, we make the following contributions,
(1) We propose a self-supervised model-to-person VTON solution that shows significant performance improvement over the previous methods, especially for complex human arm poses (see Fig.~\ref{fig: demo}). (2) The proposed part-by-part warping technique works well in computing the target warps requiring overlap among different parts of it. (3) We discuss the limitations of TPS transform in the current problem context and propose a hand-crafted feature based warping method that is consistent with the human body geometry. (4) We propose a mask prediction network (MPN) that aids in identifying the occluded regions of the target clothing and thereby guides our proposed image synthesizer network (ISN) to fill respective clothing details in those regions.

Note that this work focuses on upper body clothing only; however, can be extended for lower body clothing too. In the rest of the paper, we present a brief literature survey in Sec.~\ref{related_works}, solution approach in Sec.~\ref{methodology}, and, the experiments in Sec.~\ref{experiments}. Finally, we conclude in Sec.~\ref{sec: conclusion}. Code will be uploaded after acceptance.

\section{Related Works}
\label{related_works} 
Image-based virtual try-on approaches can be categorized into warping-based and appearance flow based approaches. Among the warping-based approaches~\cite{viton, cpvton, vtnfp, lgvton, ivcnz_roy,wang2018non, cvpr_2020_2, multiposevton} the popular warping function is Thin Plate Spline transformation function. 

The deep neural network related warping-based approaches~\cite{cpvton, vtnfp, lgvton, ivcnz_roy,wang2018non, cvpr_2020_2, multiposevton} learns the features from the clothing and the reference person images and correlates them. The features learnt from that correlation map is used to predict the parameters of TPS. The network employing this is called geometric matching network (GMN). CP-VTON~\cite{cpvton} first employed GMN and showed improvement over shape context warp idea of VITON~\cite{viton}.  VTNFP~\cite{vtnfp} additionally incorporates non-local mechanism~\cite{wang2018non} in the feature extraction part of the GMN. However, high flexibility of TPS, often causes GMN to produce undesirable warping results in presence of complex patterns in clothes. ACGPN~\cite{cvpr_2020_2} proposed to employ a second-order difference constraint to control the grid deformations causing undesirable warps.  LGVTON~\cite{lgvton} proposed a landmark guided VTON approach which used landmark based image registration approach to estimate $\theta_{TPS}$. Roy et al.~\cite{ivcnz_roy} employs GMN but learns the correspondences of features of the densepose representations~\cite{densepose} of the model and the person instead of directly learning from the images.  M2E-TON~\cite{m2e} also used densepose representations but used the idea of barycentric coordinate based texture transfer from model to person. Some GAN based image synthesis approach for attribute manipulation is proposed in~\cite{yildirim2019generating, men2020controllable, lewis2021tryongan}. Recently, Lewis et al.~\cite{lewis2021tryongan} proposed a styleGAN based novel approach of try-on on unpaired training data. For a given model and person pair it finds the optimal interpolation coefficients per layer to generate a try-on output.  GarmentGAN~\cite{garmentgan} employs two separate GANs for shape generation and appearance generation to predict the final try-on output. SieveNet\cite{sievenet} introduces a duelling triplet loss to improve the quality of the try-on output. PASTA-GAN~\cite{xie2021towards} proposed a patch-routed disentanglement module that disentangles style and spatial information in a garment. It's key contribution is employing source and target person's pose-based keypoints to decouple a garment into normalized patches and then reconstructing them to the warped garment. While we also employ pose key points but the idea of our geometric approach of warping is quite different from their approach.

Other than warping, appearance flow based approaches are also proposed~\cite{chopra2021zflow, ge2021parser, han2019clothflow, he2022style,liu2019liquid}. The very first work in this direction is proposed in~\cite{han2019clothflow}, which predicts the clothing flow from the source to target images. The main idea of appearance flow is to predict the sampling grid for clothes warping. A styleGAN~\cite{karras2019style} based appearance flow estimation method is proposed in~\cite{he2022style} which uses the style vector to capture the global context of the image. This attempts to overcomes the challenges due to considering only local appearance flow~\cite{chopra2021zflow, ge2021parser, han2019clothflow}.

Summarizing the contributions of the above methods, we see the above methods has contributed to improve the photo realism of the output. Also some method shows ability to handle complex hand postures. But most of the methods struggle in case of warping the sleeves of the long-sleeved clothing in case of significant bending of arms. The contribution of this paper targets this limitation of others. Our method works on variety of human poses, irrespective of human body shape, outfit shapes and also attempts to retain the exact texture and colors of the source clothing in the target output.

\begin{figure}[h]
	\centering
	\includegraphics[width=0.48\textwidth]{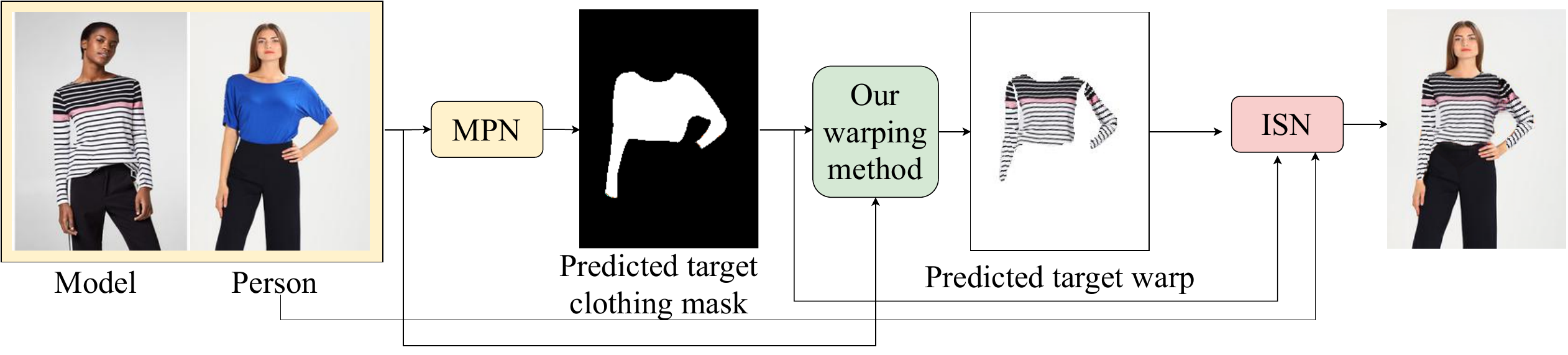}
	\captionof{figure}{Illustration of our overall virtual try-on approach.}
	\label{fig: overall_blockdia}
\end{figure}
\section{Approach}
\label{methodology}

\subsection{Overview}
Given a model's image $M$ and a person's image $P$, we propose, the problem of virtual try-on (VTON) as, $P' = \mathcal{F} (M, P)$, where $P'$ refers to the image of the person wearing the reference clothing of $M$. We compute $P'$ in mainly three steps. First, We employ a mask predictor network (MPN) (Sec.~\ref{sec_mpn}) that predicts the target clothing mask, denoting the trialed clothing region of the expected try-on output. Second, we extract the model's outfit $c$ using a human parsing method~\cite{lipssl}\footnote{Human parsing is the task of segmenting a human image into different fine-grained semantic parts e.g., head, torso, arms, legs, upper-clothes, etc.} and compute the target warp $c'$ that fits the person $P$, using our proposed warping method (Sec.~\ref{sec_warp}).
Third, we propose a novel synthesis step employing an image synthesizer network (ISN) that seamlessly combines $c'$ with $P$ to generate the final try-on output $P'$ (Sec.~\ref{sec_isn}). 
Below we discuss all the steps in detail. A simplistic block diagram demonstrating the different steps of our approach is shown in Fig.~\ref{fig: overall_blockdia}.
\subsection{Predicting the target clothing mask}
\label{sec_mpn}
This step predicts the target clothing mask serving two purposes in the subsequent stages of this work: 
(1)~it aids the synthesis module to identify the occluded regions of the source clothing which needs to be interpolated to produce a seamless try-on output,
(2)~it helps our proposed warping method in reducing the time of computation of the target warp corresponding to the sleeves of the source clothing. This will be elaborated on later during the discussion of the warping method.

To predict the mask, we train a convolutional neural network (CNN), named the mask prediction network (MPN), with the following inputs - (1)~the underclothing body shape of the model and the person, encoded using their densepose representations~\cite{densepose}~\footnote{A densepose representation maps all human pixels of an RGB image, to the 3D surface of the human body, thus providing a precise estimate of the human body shape under the clothing.}. In addition, we also provide the face and head segments of both the model and the person extracted using~\cite{lipssl}. (2)~the mask of $c$, and (3)~semantically segmented human parts~\cite{cdcl} of the model. A block diagram of MPN along with the inputs and output are demonstrated in Fig.~\ref{fig: all_1}(b).
Note that providing the face and head segments along with the densepose as input gives the whole detail of a human body. Whereas, a more refined level of details is provided by the segmented human parts.
 
The architecture of MPN is demonstrated in Fig.~\ref{fig: all_1}(b). We incorporate correlation layers in MPN that computes the linear relationship between the body shape and pose features of the model and that of the person (Observe Fig.~\ref{fig: all_1}(b)). The human parsing branch at the end of this module (demonstrated in Fig.~\ref{fig: all_1}(b)) predicts the human parsing of $P'$ given the non-target clothing details of $P$ e.g., the lower body clothing, the face and hair segments, and the predicted mask from the previous branch. This is enforcing another constrain on the main branch to predict a mask consistent with the final human parsing.

\subsection{Predicting the target warp}
\label{sec_warp}

\begin{figure*}[!htp]
	\centering
	\includegraphics[width=0.9\textwidth]{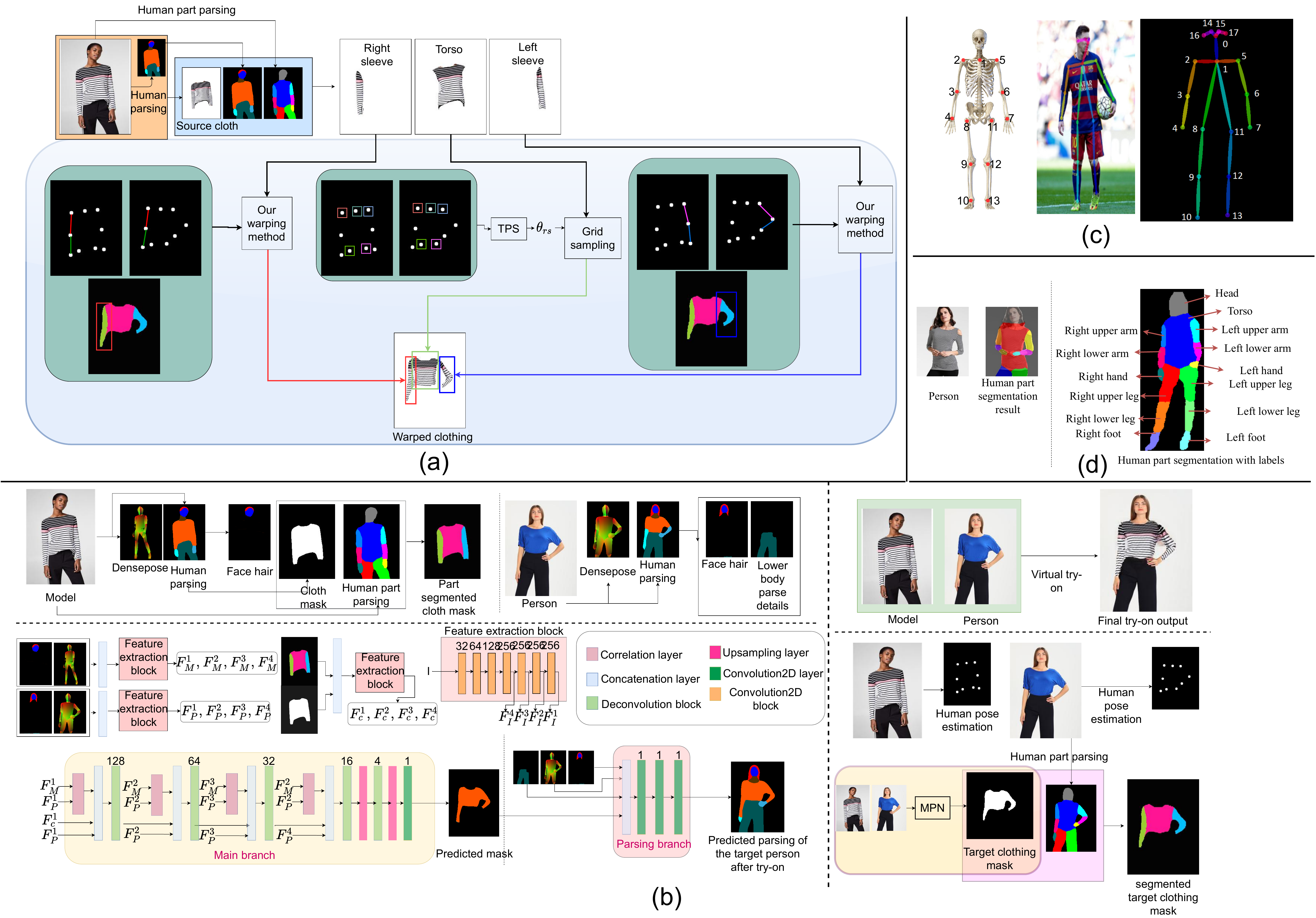} 
	\captionof{figure}{(a)~Illustration of the method for predicting the target warp, (b)~Block diagram of the proposed mask predictor network (MPN), (c)~Demonstration of human landmarks, (d)~Demonstration of human part parsing~\cite{cdcl} results and pose key points.Best viewed in electronic version.}
	\label{fig: all_1}
\end{figure*}

\begin{figure*}[ht]
	\centering
	\includegraphics[width=1\textwidth]{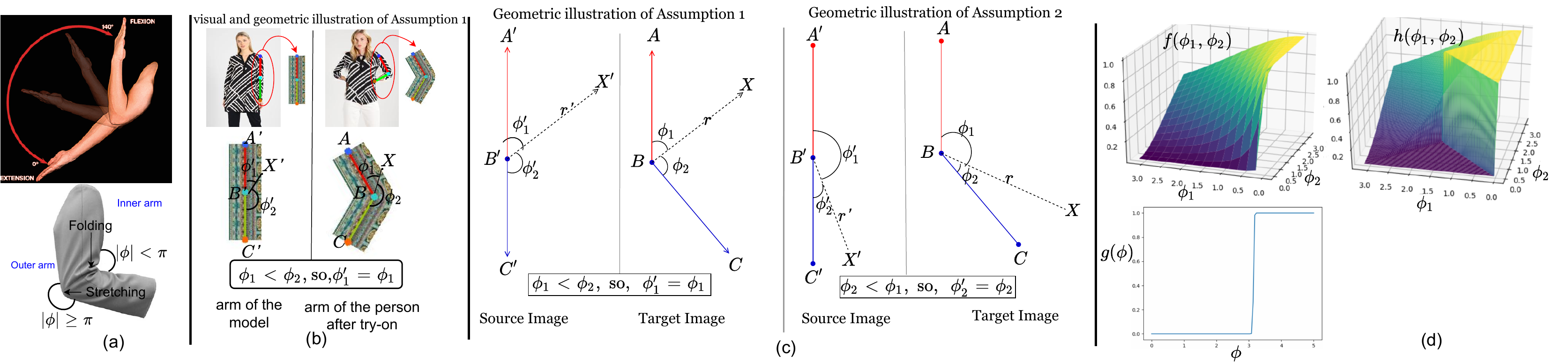}
	\captionof{figure}{(a)~(top)Elbow flexion and extension (picture courtesy~\cite{elbow_flexion}), (bottom)~stretch and folds in clothing sleeve due to elbow flexion  i.e., arm bending (picture courtesy~\cite{sleeve_folding}). (b) A graphical illustration of the arm bending phenomenon in human. A sample pair of model and person relevant to the illustrated scenario is given above for better understanding, (c)~Plot of functions $f(\phi_1, \phi_2)$, $g(\phi)$ and $h(\phi_1, \phi_2)$, (d)~Geometrical illustration of our warping method for sleeves warping for the two different scenarios. (Left)~Example of a case when assumption 1 holds. (Right)~Example of a case when assumption 2 holds. Here $\{A, B, C\}$ and $\{A', B', C'\}$ are the landmarks corresponding to the arm of the person and the model respectively. $X$ refers to the point belonging to the sleeve segment of the target warp and $X'$ is its corresponding source pixel.}
	\label{fig: all_2}
\end{figure*}

This step predicts $c'$ from $c$. We do it part-by-part, by first segmenting $c$ into 3 semantic parts - torso ($c_{torso}$), left sleeve ($c_{lsleeve}$), right sleeve ($c_{rsleeve}$) and thereafter opting skeletal structure specific warping approaches. Different parts of human body have different skeletal structures causing different limitations of movement in each part. This effects the transformation of the related clothing part when the clothing is transferred from one human's shape-pose state to another. For instance, in our case from the model's shape-pose state to that of the person's. Our idea of part-by-part warping is inspired from this fact only. Below, first we discuss our proposed method of warping the sleeves i.e., predicting $c_{lsleeve}'$, $c_{rsleeve}'$, which is one of the major contributions of this paper. Thereafter, we discuss the approach of warping the torso part of the clothing i.e., computing $c_{torso}'$. The human part segmentation method~\cite{cdcl} used to segment the clothing into different parts is portrayed in Fig.~\ref{fig: all_1}(d). 

In order to compute the target warp of the model's sleeves we employ feature correspondences from the arm of the model to that of the person. Considering features related to the arm is obvious here because sleeves relate to arm part only. Formally, we propose geometric warping approach, where for each source pixel $X' \in c_{lsleeve} (c_{rsleeve})$ our goal is to compute the corresponding target pixel $X \in c_{lsleeve}'(c_{rsleeve}')$.

In human skeleton each arm consists of 3 bones, humerus bone in the upper arm, and radius and ulna in the lower arm. Now considering radius and ulna together, an arm may be represented as consisting of 2 bones, each corresponding to the upper and lower arm connected at a the common elbow joint. Using human pose representation~\cite{openpose} we can get the location of this elbow joint and the line segments corresponding to the two bones.  For instance as shown in Fig.~\ref{fig: all_1} (c) each arm contains only 3 landmarks. The line segments connecting these landmarks for example, \{(2,3), (3,4)\} for right arm and \{(5,6), (6, 7)\} for left arm may be considered the representative of the bones. Hence, geometrically, consider the arm of the target person consists of the following line segments, \{($B$, $A$), ($B$, $C$)\} and that of the source person as \{($B'$, $A'$), ($B'$, $C'$)\}. Please see Fig.~\ref{fig: all_2}\{(b), (c)\} for visual clarifications. We employ the correspondences of these lines, for instance from ($B'$, $A'$) to ($B$, $A$) and ($B'$, $C'$) to ($B$, $C$) for computing the target sleeve warp. Note that instead of only landmarks or bone joints correspondences i.e., \{A', B', C'\} to \{A, B, C\} as proposed in~\cite{lgvton}\footnote{LGVTON considers correspondences of 3 landmarks of each arm from source to target.} our approach is more constrained as we are considering the correspondences of the human bones which is anatomically more consistent.

We use polar coordinates relative to the line segments $BA$ and $B'A'$ to define the location of $X$ and $X'$ respectively\footnote{Note that considering the locations with respect to the lines $BC$ and $B'C'$ will not effect the idea of this method}. Also polar system is useful in our case, which will becomes clear with further readings. As shown in Fig.~\ref{fig: all_2}(b) with respect to $BA$ location of $X$ is $(r, \phi_1)$ where $r = \|BX\|$ and $\phi_1$ = $\angle{XBA}$. Similarly with respect to $B'A'$ the location of $X'$ is $(r', \phi_1')$. Since we already have the locations of landmarks $A, B, C$, and, $A', B', C'$, so for each $X$ we have the values of $(r, \phi_1)$. Now, with regard to the backward mapping concept for each of $X$ our goal is to compute the corresponding $X'$ i.e., the values of $(r', \phi_1')$. Note that here we compute $X'$ only for the $X \in \{c_{sleeve}' \cap \mathcal{S}\}$, where $\mathcal{S}$ is the target clothing mask predicted by MPN, and, $c_{sleeve}' = (c_{lsleeve}' \cup c_{rsleeve}')$. This aids in reducing the computation time.

Below, we first discuss the method to compute the angular coordinate $\phi_1'$ followed by discussing our idea of computing the radial coordinate $r'$. Now before continuing further, let us introduce some additional notations. Let $\phi_2$ = $\angle{CBX}$, $\phi_1'$ = $\angle{X'B'A'}$. $\phi_2'$ = $\angle{C'B'X'}$, $\phi$ = $\angle{CBA}$ = $\phi_1 + \phi_2$.

In general human arm undergoes bending which is called \textit{elbow flexion}. The bending amount is quantified using the \textit{flexion angle} ($\in [ 0^{\circ} \text{(no bending)}, 145^{\circ} \text{(max bending)}]$)~\footnote{https://en.wikipedia.org/wiki/Elbow}(ref to Fig.~\ref{fig: all_2}(a)). Due to bending the sleeve undergoes folding in the inner part and stretching in the outer part (Fig.~\ref{fig: all_2}(a)). We observed, due to bending the relative angular position of a source pixel $X'$ with its closest line remains unchanged in the target warp. For instance, as shown in Fig.~\ref{fig: all_2}(b), X is closer to $BA$ than $BC$ as $\phi_1 < \phi_2$ so $\phi_1' = \phi_1$. We include this observation in the form of two assumptions,
\begin{itemize}
	\item Assumption 1 - when $X$ is closer to the line $BA$ i.e., when $\phi_1 < \phi_2$,
	the relative angular position of $X$ with respect to line $BA$  i.e., $\phi_1$, will be equal to the angular position of $X'$ with respect to line $B'A'$ which is $\phi_1'$, i.e., $\phi_1' = \phi_1$.
	\item Assumption 2 - In a similar sense, when $X$ is closer to the line $BC$ i.e., when $\phi_2 < \phi_1$, then $\phi_2' = \phi_2$. Note that saying $\phi_2' = \phi_2$ is equivalent to saying $\phi_1' = \phi' - \phi_2$.
\end{itemize}

In the elbow region bending causes 2 kinds of warping of the sleeves, folding and stretching as shown in Fig.~\ref{fig: all_2}(a). In the outer arm area near the elbow region where generally $\phi_1$ and $\phi_2$ are very close, strictly adhering to one of two assumptions does not produce realistic result. Instead, then we compute $\phi_1'$ as the weighted combinations of the values of $\phi_1'$ due to the two assumptions; where the weight is computed by a function $f(\cdot, \cdot)$. Whereas, as $X$ moves towards the upper and lower end of the arm either of the two assumptions can be used to compute $\phi_1'$. Based on this we compute $\phi_1'$ for the pixels in the outer arm region as,
\begin{equation}
	\phi_1' = \phi_1 (1 - f(\phi_1, \phi_2)) + (\phi' - \phi_2) f(\phi_1, \phi_2),
	\label{eq: theta1prime}
\end{equation}
where, 
\begin{equation}
    f(\phi_1, \phi_2) = \frac{\phi_1^2}{\phi_1^2 + \phi_2^2}
    \label{eq: func_f}
\end{equation}
The choice of this function is intuitive. Observe in Fig.~\ref{fig: all_2}(d) that the functional value of $f$ takes smooth transition from 0 to 1 from the region $\phi_1 < \phi_2$ (criterion of assumption 1) to $\phi_1 > \phi_2$ (criterion of assumption 2). This smooth transition ensures that near the elbow in the outer arm area i.e., when $\phi_1$ is close to $\phi_2$ a weighted a combination of the effects of both the assumptions hold for stretching unlike the outer arm.

For the sleeve area in the inner arm region either of the two assumptions holds. A rounded version of the function $f$ may be used in such case, because rounding removes the effect of smooth transition which is not required here. So for the pixels in the inner arm region we compute $\phi_1'$ as, 
\begin{equation}
	\phi_1' = \phi_1 (1 - Round(f(\phi_1, \phi_2))) + (\phi' - \phi_2) Round(f(\phi_1, \phi_2)).
	\label{eq: theta1prime}
\end{equation}

To combine these two cases in one expression we need a function to distinctly identify whether a pixel belongs to the inner arm or outer arm. Notice in the inner arm $\phi < \pi$ and in the outer arm $\phi \geq \pi$ (Fig.~\ref{fig: all_2}(a)). Based on this we define $g(\cdot, \cdot)$ as a logistic function that varies from 0 to 1 near $\pi$. The idea of taking a logistic instead of a step function here is to make a smooth transition between the regions $\phi < \pi$ and $\phi > \pi$ so that a realistic warping result is obtained. The functional form of $g$ is,
\begin{equation}
    g(\phi) = \frac{1}{1 + e^{a(\pi - \phi)}}.
    \label{eq: func_g}
\end{equation}
Observe, when $\phi < \pi$ (inner arm), $g(\cdot, \cdot) \to 1$, and, when $\phi > \pi$ (outer arm), $g(\cdot, \cdot) \to 0$. Now we know, in the inner arm $f(\phi_1, \phi_2)$ holds and in the outer arm $Round(f(\phi_1, \phi_2))$. Combining these two using $g(\cdot, \cdot)$ we get $h(\cdot, \cdot)$ as follows, 
\begin{equation}
	h(\phi_1, \phi_2) = g(\phi) f(\phi_1, \phi_2) +  (1 - g(\phi))Round(f(\phi_1, \phi_2)). 
	\label{eq: func_h}
\end{equation}

Hence, for all pixel $X$, we may computer $\phi_1'$  as,
\begin{equation}
	\phi_1' = \phi_1 (1 - h(\phi_1, \phi_2)) + (\phi' - \phi_2) h(\phi_1, \phi_2),
	\label{eq: theta1prime}
\end{equation}

\begin{figure}[h]
	\centering
	\includegraphics[width=0.2\textwidth]{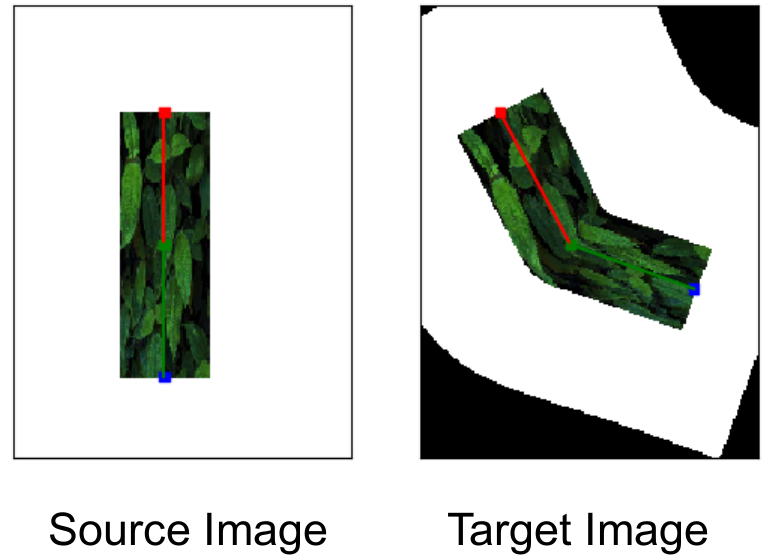}
	\captionof{figure}{Our result depicting the effect of scaling of lines in our method. As the length of the line increases from source to target the corresponding part of the rectangle looks zoomed-in in the result.}
	\label{fig: warp_demo2}
\end{figure}

Now that, we have computed $\phi_1'$, we need to compute $r'$ as previously mentioned. Before we compute $r'$ let us understand, a point closer to $B'C'$ (or $B'A'$) should be proportionally similar in distance to $BC$ (respectively $BA$). To maintain this scaling effect we multiply $r$ with the ratio of the source and the corresponding target line i.e., $\|B'C'\|$ and $\|BC\|$  (resp. $\|B'A'\|$ and $\|BA\|$). For example, if $\|B'A'\| < \|BA\|$ then, $r'< r$, as $\frac{\|B'A'\|}{\|BA\|} < 1$. A demonstration of this case is given in Fig.~\ref{fig: warp_demo2}. In this example, the line $BA$ (red-colored) is longer than $B'A'$, hence, the upper part of the sleeve (rectangle) in the transformed image is scaled accordingly. 

Based on this concept, we obtain the value of $r'$ using the following formula. 
\begin{equation}
	r' = r\left\{(1 - h(\phi_1, \phi_2)) \frac{\|B'A'\|}{\|BA\|}  +  h(\phi_1, \phi_2) \frac{\|B'C'\|}{\|BC\|}\right\}.
	\label{eq: rprime}
\end{equation}

Now, we have computed the radial and angular coordinates $r'$ and $\phi_1'$ of $X'$ respectively and, we already have the coordinates of $A'$ and $B'$. Therefore, computing the location of $X'$ in Cartesian is straightforward. Hence, we do not describe it in detail. Note that we compute the source pixels corresponding to only the pixels of the target clothing regions predicted by the proposed MPN.

The procedure of sleeve warping is formally presented in Algorithm.~\ref{algo: 1}. In order to keep the main manuscript concise and to the point only some further analysis on how our method handles occluded areas in the source clothing regions is explained in the supplementary materials.

\begin{algorithm}
\caption{Algorithm for computing the source pixel $X' \in c_{lsleeve}(c_{rsleeve})$ corresponding to a location $X \in c_{lsleeve}'(c_{rsleeve}')$. See Fig.~\ref{fig: all_2}(b) for a visual understanding of the notations.}
\label{algo: 1}
\begin{algorithmic}[1]
\renewcommand{\algorithmicrequire}{\textbf{Input: \{$A', B', C'$\},\{$A, B, C$\}. An arbitrary pixel $X$}}
\renewcommand{\algorithmicensure}{\textbf{Output: Pixel $X'$ corresponding to $X$}}
\REQUIRE
\ENSURE
\STATE Compute the radial coordinate $r$ and angular coordinate $\phi_1$ of $X$ as $r = \|BX\|$ and $\phi_1 = \angle{XBA}$
\STATE Given $A, B, C$ compute $\phi_1, \phi_2$; \\where $\phi_1$ = $\angle{XBA}$ and $\phi_2$ = $\angle{CBX}$.
\STATE Compute $f(\phi_1, \phi_2)$ using Eq.~\ref{eq: func_f}, and $g(\phi)$ using Eq.~\ref{eq: func_g} where $\phi = \phi_1 + \phi_2$
\STATE Compute $h(\phi_1, \phi_2)$ from $f(\cdot, \cdot)$ and $g(\cdot)$ using Eq.~\ref{eq: func_h}
\STATE Compute the angular coordinate $\phi_1'$ of $X'$ using Eq.~\ref{eq: theta1prime}
\STATE Compute the radial coordinate $r'$ of $X'$ using the Eq.\ref{eq: rprime}
\STATE Now we have the radial and angular coordinates i.e., $r'$ and $\phi_1'$ of $X'$ respectively. Therefore, computing the location of $X'$ in Cartesian coordinate system is straightforward as we already have the coordinates of $A'$ and $B'$.
\RETURN $X'$ 
\end{algorithmic}
\end{algorithm}

To compute $c_{torso}'$ from $c_{torso}$ we adopt human landmark correspondences to compute TPS transform~\cite{lgvton}. 

But in this application, since we are interested in the target clothing region only, so, we compute the source pixels corresponding to the pixels belonging only to the target clothing region predicted by the proposed MPN
 
\subsection{Try-on image synthesis}
\label{sec_isn}
The objective of this stage is to combine $c'$ with $P$ to generate a plausible and photo-realistic try-on result. Our warping method computes $c'$ from only the visible clothing areas of the model's clothing. But due to the difference in pose between model and person some occluded areas of the model's clothing might become visible in the target. Also, long hairs often occlude clothing areas. Considering the target clothing mask predicted by MPN as $\mathcal{S}$, and the mask of the predicted warp $c'$ as $c'_m$, we obtain such areas as, $\mathcal{S}\cdot \bar{c'_m}$, where $\cdot$ denotes dot product and $\bar{(.)}$ denotes complement. The main goals of ISN are to inpaint this region and produce a seamless combination of $c'$ and $P$, while also removing the previous clothing details from $P$. 

To achieve the above objectives we trained an encoder-decoder based convolutional neural network (CNN). We call it image synthesizer network (ISN). The inputs to this network are: (i)~the predicted target clothing mask from the MPN, (ii)~the body shape of the target person represented by its densepose representation, (iii)~
instead of providing the predicted target warp and the person image separately, we provide a combined representation of them. To compute this, the upper body details i.e., previous clothing (e.g., t-shirt, etc.), exposed skin areas are masked in the person image, and the predicted warp is overlaid on it. (iv)~ The inpainting mask $\bar{\mathcal{S}} \cdot \bar{c'_m}$ (0 denotes the pixel with missing details).

We trained this network using self-supervision employing inpainting related loss functions as suggested in~\cite{liu2018image}. 

\section{Experiments}
\label{experiments}
\textbf{Dataset.} 
We experiment on MPV dataset~\cite{multiposevton} which is a benchmark dataset in current problem domain. It contains 35,687 images of fashion models captured from various angles with their upper-body clothing images given separately that in total is 13,524. We randomly collected a test set with 3000 image pairs. Note that this dataset does not contain any image of two different models wearing same garment. Therefore, our test set does not have any ground truth images. The training set contains 20,034 same person image pairs. We chose MPV over the other benchamrk datasets DeepFashion~\cite{deepfashion} and VITON~\cite{viton}, because DeepFashion~\cite{deepfashion} does not contain the separate garment images. Also MPV and VITON both are collected from same source website so have similar data and even overlap in contents.

\subsection{Quantitative analysis}
We quantitatively evaluate our results on two metrics - \textit{Fréchet Inception Distance (FID)}~\cite{fid}, the structural similarity metric (SSIM). FID measures the similarity between two sets of images based on the features extracted from a layer of inception v3 model~\cite{inceptionv3} pretrained on ImageNet~\cite{imagenet}. 
A lower value of FID indicates better results. \textit{SSIM~\cite{ssim}} is an image similarity metric, ranges in [0, 1] (1 indicates same images).

We compare our performance with, CP-VTON~\cite{cpvton}, MGVTON~\cite{multiposevton}, LGVTON~\cite{lgvton}~\footnote{Note that the results of LGVTON are reported based on human landmark annotations only because fashion landmarks are not available in the MPV dataset.}, Roy et al.~\cite{ivcnz_roy}, ACGPN~\cite{cvpr_2020_2} and He et al.~\cite{he2022style} and report the SSIM and FID scores in Table.~\ref{results_quantitative} (SSIM score is reported only for the datasets containing ground truth). Comparison with some other benchmarks PFAFN~\cite{ge2021parser} and PASTA-GAN~\cite{xie2021towards} are presented in Table.~\ref{results_quantitative_pasta_gan}. Note that due to difficulty in execution of PASTA-GAN's official implementation we provide our method's FID score on PASTA-GAN's MPV test set (provided by the authors). Note that in order to make fare comparison we compare with almost equal number of M2P and C2P methods. M2P methods - LGVTON, Roy et al., PASTA-GAN, PFAFN and C2P methods - MGVTON, CP-VTON, ACGPN, He et al. Also like ours, both LGVTON and PASTA-GAN uses model and person's pose key points in computing the target warp.

The C2P methods~\cite{multiposevton, cpvton, cvpr_2020_2, he2022style} take separate garment image as input. Such garments are kept in one pose only and also have no occluded parts. But in case of model image the clothing might be occluded and also pose variability between the source and the target person might be there. To make fair comparison we execute these methods on both clean garment image and segmented garment image. Results in Table.~\ref{results_quantitative} shows that our method outperforms the most of the other methods in terms of both metrics in both the input setting. Though He et al. achieves better score in case of no occlusion and no source clothing's pose variability. It is also observable in Table.~\ref{results_quantitative_pasta_gan} that also on the MPV test set of PASTA-GAN our method scores better compared to others. 

\begin{table}[!h]
\centering
	\caption{Quantitative evaluation on different datasets. In column 2 the inputs of MGVTON, CP-VTON, ACGPN, and He et al. is ($c$, $P$) and that of the others is ($M$, $P$). In column 5, MGVTON, CP-VTON, ACGPN, and He et al. are run in their default inputs setting i.e., is ($C$, $P$). 
	Since in the training set source and the target images are same, therefore no variation in pose, so, no question of occlusion, hence result will be same for both the clothing sources.
	}
\begin{tabular}{c|c|cc|c}
\toprule
& \multicolumn{1}{c}{Test Set}  & \multicolumn{2}{c}{Training Set} & \multicolumn{1}{c}{Test Set}\\

& \multicolumn{1}{c}{(M2P)} & \multicolumn{2}{c}{} & \multicolumn{1}{c}{(C2P)}\\

Method & FID$\downarrow$ & FID$\downarrow$ & SSIM$\uparrow$ & FID$\downarrow$\\
 \midrule
MGVTON(ICCV, 19) & 47.23 & 35.70 & 0.76 & 38.19\\
CP-VTON (ECCV, 18) & 50.67 & 21.03 & 0.74 & 40.02\\
LGVTON(MTA, 22) & 28.86 & 12.06 & 0.89 & - \\

Roy et al.(IVCNZ, 2020) & 28.41 & 14.34 & 0.80 & - \\

ACGPN (CVPR, 20) & 27.80 & 14.35 & 0.88 & 17.87\\
He et al.(CVPR, 22) & 18.31 & 11.45 & 0.89 & 9.38\\
Ours & \textbf{16.38} & \textbf{9.45} & \textbf{0.93} & - \\
\bottomrule
\end{tabular}
\label{results_quantitative}
\end{table}

\begin{table}[!h]
\centering
	\caption{The FID scores on the MPV test set used in PASTA-GAN. The scores of all the methods except ours are taken from the paper PASTA-GAN.
	}
\begin{tabular}{c|c}
\toprule
& \multicolumn{1}{c}{Test Set}  \\

& \multicolumn{1}{c}{(M2P)}\\

Method & FID$\downarrow$ \\
 \midrule
CP-VTON (ECCV, 18) & 37.72\\
ACGPN (CVPR, 20)& 23.20\\
PFAFN (CVPR, 21)& 17.40\\
PASTA-GAN (NeurlIPS, 21) & 16.48\\
Ours & \textbf{15.74} \\
\bottomrule
\end{tabular}
\label{results_quantitative_pasta_gan}
\end{table}

\begin{figure*}[!h]
	\centering
	\includegraphics[width=0.8\textwidth]{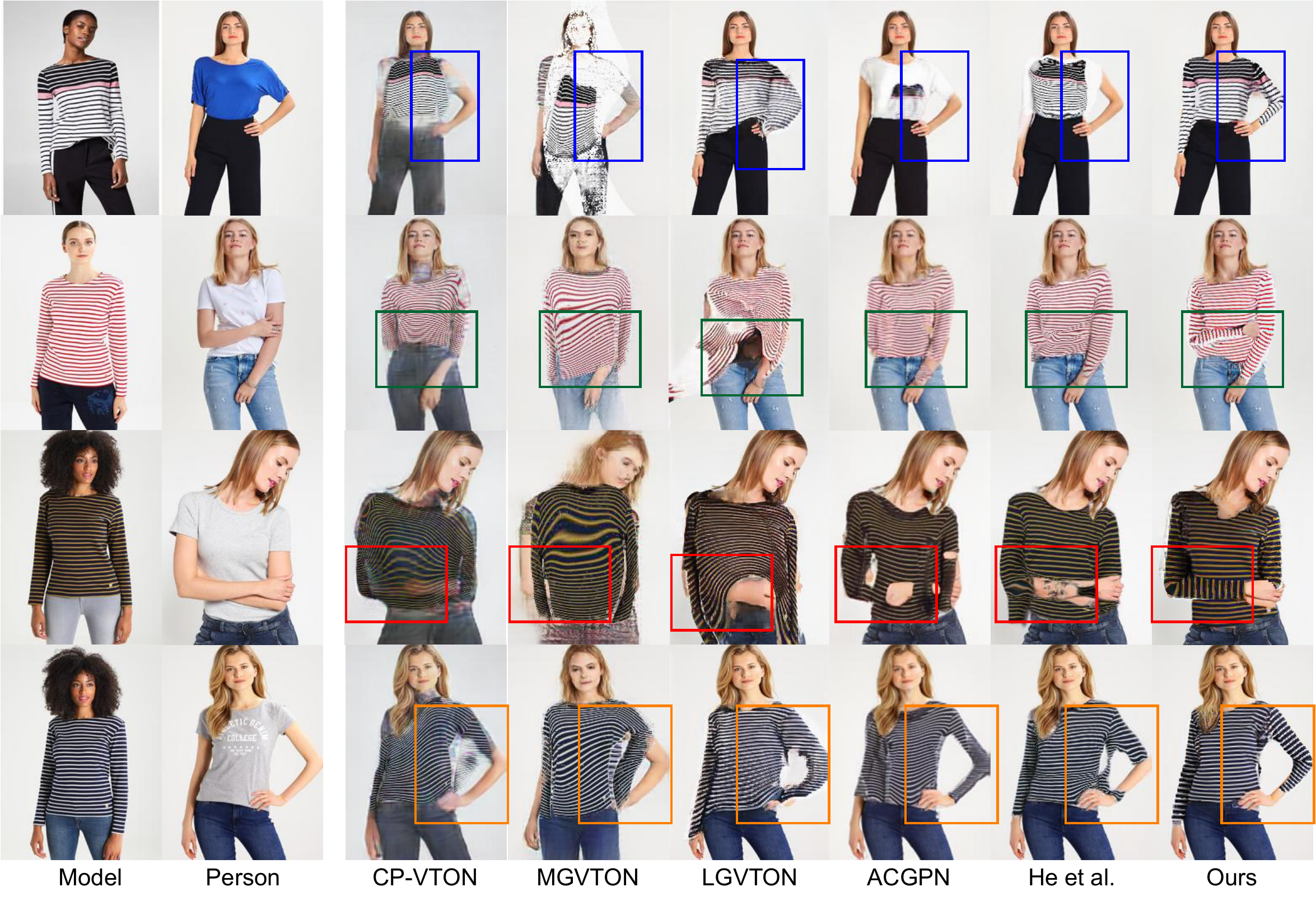}
	\captionof{figure}{Comparative study on the cases of significant arm bending.}
	\label{fig: results_compare1}
\end{figure*}
\begin{figure}[!h]
	\centering
	\includegraphics[width=0.4\textwidth]{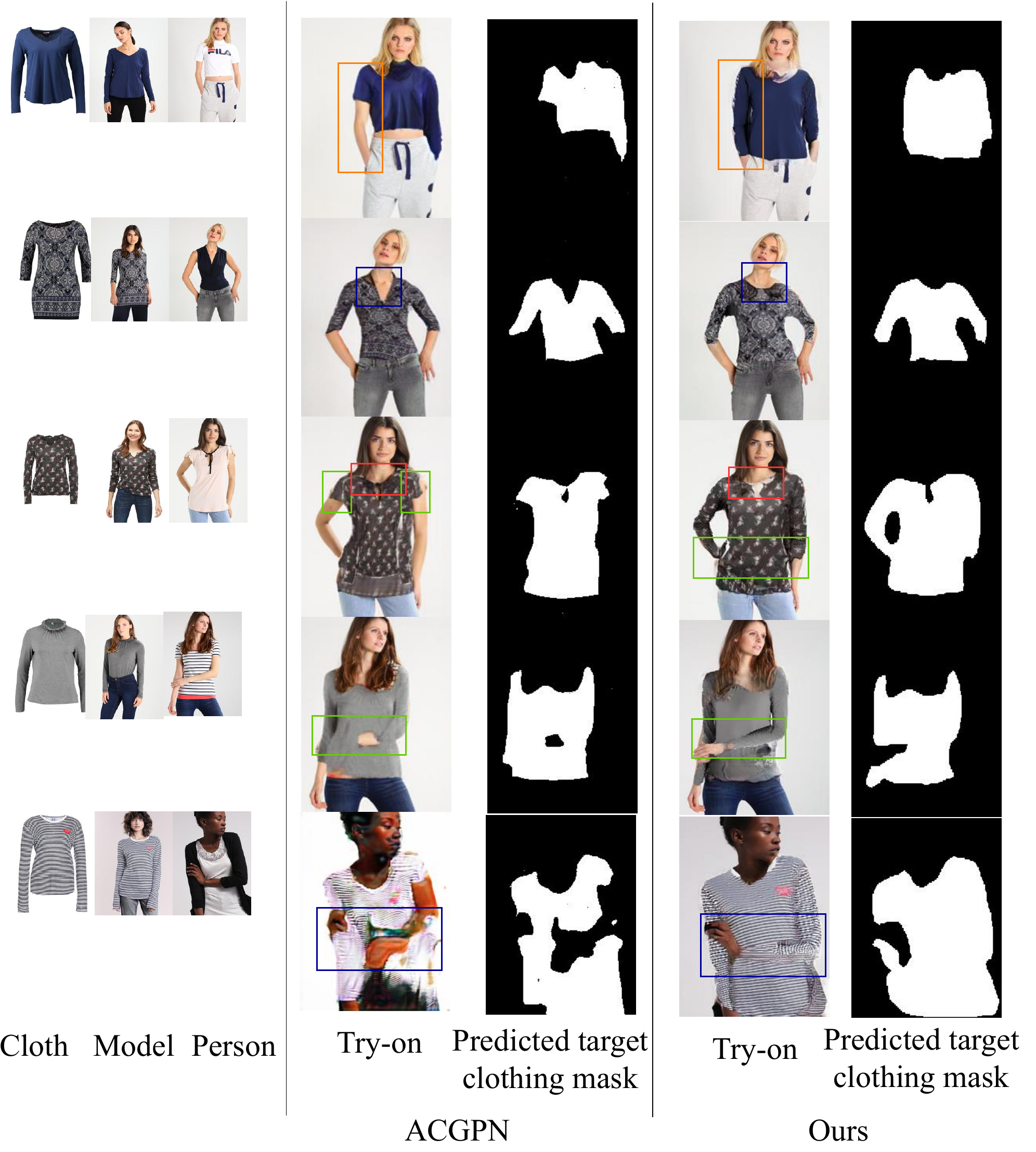}
	\captionof{figure}{Qualitative comparison of the proposed MPN with the semantic generation module (SGM) of ACGPN. We have highlighted the areas to be noticed in each of the results. Observe in most of the cases SGM does not preserve the features of the source clothing in the try-on result, instead keeps those features same as the old cloth of the person (i.e., before try-on).}
	\label{fig: mpn_output1}
\end{figure}

\subsection{Qualitative Analysis}
Some qualitative comparisons are given in Fig.~\ref{fig: results_compare1}. Note that the limitations of the methods when the segmented clothing image is given as input can be verified from the quantitative analysis done before. Here we chose to compare in every method's default input setting i.e, the garment image as input for~\cite{cvpr_2020_2, multiposevton, cpvton, he2022style}
and the model image to~\cite{lgvton}. This is to compare all the methods at the best of their ability.

It can be observed that while handling critical hand bending all the methods except ours fails to perform well. In case of CP-VTON and MGVTON probably the poor result occurs due to impaired learning of human body geometry by the geometric matching network. ACGPN shows some improvement due to the second-order difference constraint. 
Notice that CP-VTON and ACGPN have tried to compensate for the limitations of their warping method in their try-on stage. For instance, the blurry areas in their predicted warps which are mostly because those regions are being filled by the try-on synthesis module of the respective methods which are not strong enough to fill the texture or color accurately. In comparison to others, our results show better-predicted warps. Similar to LGVTON employing the idea of landmarks correspondence in warping produces correct deformation in the torso region. However, unlike the others, we can handle the bending of the sleeves due to our geometric warping approach; and also our part-by-part warping approach enables us in handling the cases of overlap between sleeve and torso.
We also present a visual comparison among the results of our MPN and the corresponding semantic generation module (SGM) of ACGPN in Fig.~\ref{fig: mpn_output1} because the objectives of these modules are same. The results show some significant features (e.g., neck pattern, sleeve length, highlighted in Fig.~\ref{fig: mpn_output1}) of the target clothing mask are predicted better by our MPN compared to SGM.

\subsection{Ablation Study}
\label{ablation_study}
We analyze the significance of employing MPN and its human parsing branch in our approach. In the former case, we compute the warping results without using the mask predicted by MPN $\mathcal{S}$, i.e., we only compute $X'$ corresponding to the $X \in C_{sleeve}'$ instead of $X \in \{C_{sleeve}' \cap \mathcal{S}\}$.

\begin{figure}[!t]
	\centering
	\includegraphics[width=0.5\textwidth]{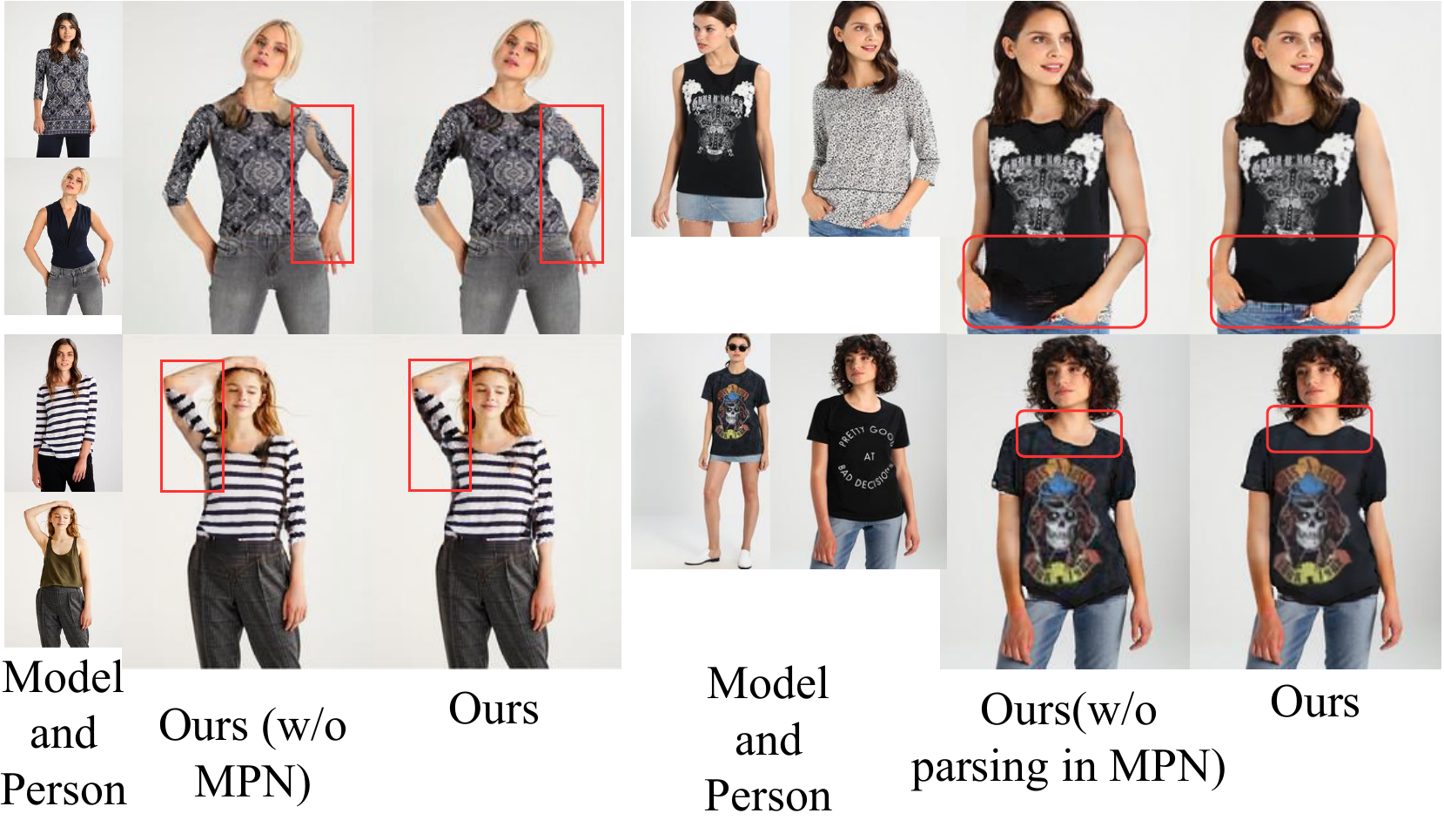}
	\captionof{figure}{illustration of the significance of MPN and MPN without parsing branch in the synthesis stage.}
	\label{fig: ablation2}
\end{figure}

For the latter case, we trained an instance of MPN without the parsing branch and compute the final results using that. Compared to the w/o MPN instance, our method secures better score, which can also be verified from the visual comparison given in Fig.~\ref{fig: ablation2}, e.g., see the previously occluded parts are being inpainted in target in our results. Compared to the without parsing instance, our results secure a better FID score as shown in Table.~\ref{tab: ablation2}. A visual comparison of the results are portrayed in Fig.~\ref{fig: ablation2}, shows improvement in some of the predicted feature details e.g., near the collar and in the adjacent region of upper and lower body clothes we see improved details.

\begin{table}[h]
	\centering
	\caption{On the significance of MPN and MPN without parsing branch in the synthesis stage of our method.}
	\label{tab: ablation2}
	\begin{tabular}{p{6cm}p{1cm}}
		\toprule
		Method & FID$\downarrow$\\
		\hline
		Ours (target warping w/o mask from MPN) & 16.69\\
		Ours (w/o parsing branch in MPN) & 16.66\\
		Ours & 16.38 \\
		\bottomrule
	\end{tabular}%
\end{table}

\section{Conclusion}
\label{sec: conclusion}

This work proposes a novel virtual try-on approach. Our method compared to the previous approaches shows improvement in terms of handling complex postures of the input person images. In the upper body of human the torso undergoes limited deformation but human arms can move in variety of ways. During try-on this might cause the target warp to have significant deformation. Most of the previous methods employed one transformation function to compute the target warp. But different parts should be warped independently to model the cases of overlap and also due to the different natures of movement of different parts of the body. To address this, we follow a part-based approach. We propose a geometric feature-based warping method to warp the sleeve that overcomes several issues of previous methods. Our warping method is guided by pose key points and follows constrains of human body movements. We also propose two learning-based modules that aids in computing the warp more efficiently and synthesizes a seamless output while taking care of the occluded source clothing regions. 
While this method shows improvement in performance but the the computation of landmarks and parsing adds overhead. Considering the utility of the proposed geometric features in future we plan to work towards exploring its possibility towards guiding appearance flow based methods.
\bibliographystyle{IEEEtran}
\bibliography{bibliography}
\end{document}